\pdfoutput=1

\documentclass[11pt]{article}

\usepackage{ACL2023}

\usepackage{times}
\usepackage{latexsym}

\usepackage[T1]{fontenc}

\usepackage[utf8]{inputenc}

\usepackage{microtype}

\usepackage{inconsolata}

\usepackage{graphicx}
\usepackage{amsmath}
\usepackage{amsfonts}       
\usepackage{multirow}

\title{Domain-Agnostic Neural Architecture for Class Incremental Continual Learning in Document Processing Platform} 

\author{
    Mateusz Wójcik$^{1, 2}$, Witold Kościukiewicz$^{1, 2}$, Mateusz Baran$^{1, 2}$, \\ {\bf Tomasz Kajdanowicz$^{1}$,} {\bf Adam Gonczarek$^{2}$} \\
    $^1$Wroclaw University of Science and Technology $^2$Alphamoon Ltd., Wrocław \\
    \texttt{\{mateusz.wojcik,tomasz.kajdanowicz\}@pwr.edu.pl} \\
    \texttt{adam.gonczarek@alphamoon.ai}
}

\begin{document}
\maketitle
\begin{abstract}
Production deployments in complex systems require ML architectures to be highly efficient and usable against multiple tasks. 
Particularly demanding are classification problems in which data arrives in a streaming fashion and each class is presented separately.
Recent methods with stochastic gradient learning have been shown to struggle in such setups or have limitations like memory buffers, and being restricted to specific domains that disable its usage in real-world scenarios.
For this reason, we present a fully differentiable architecture based on the Mixture of Experts model, that enables the training of high-performance classifiers when examples from each class are presented separately. We conducted exhaustive experiments that proved its applicability in various domains and ability to learn online in production environments. The proposed technique achieves SOTA results without a memory buffer and clearly outperforms the reference methods.

\end{abstract}

\section{Introduction}

\begin{figure}
\centering
  \includegraphics[width=\linewidth]{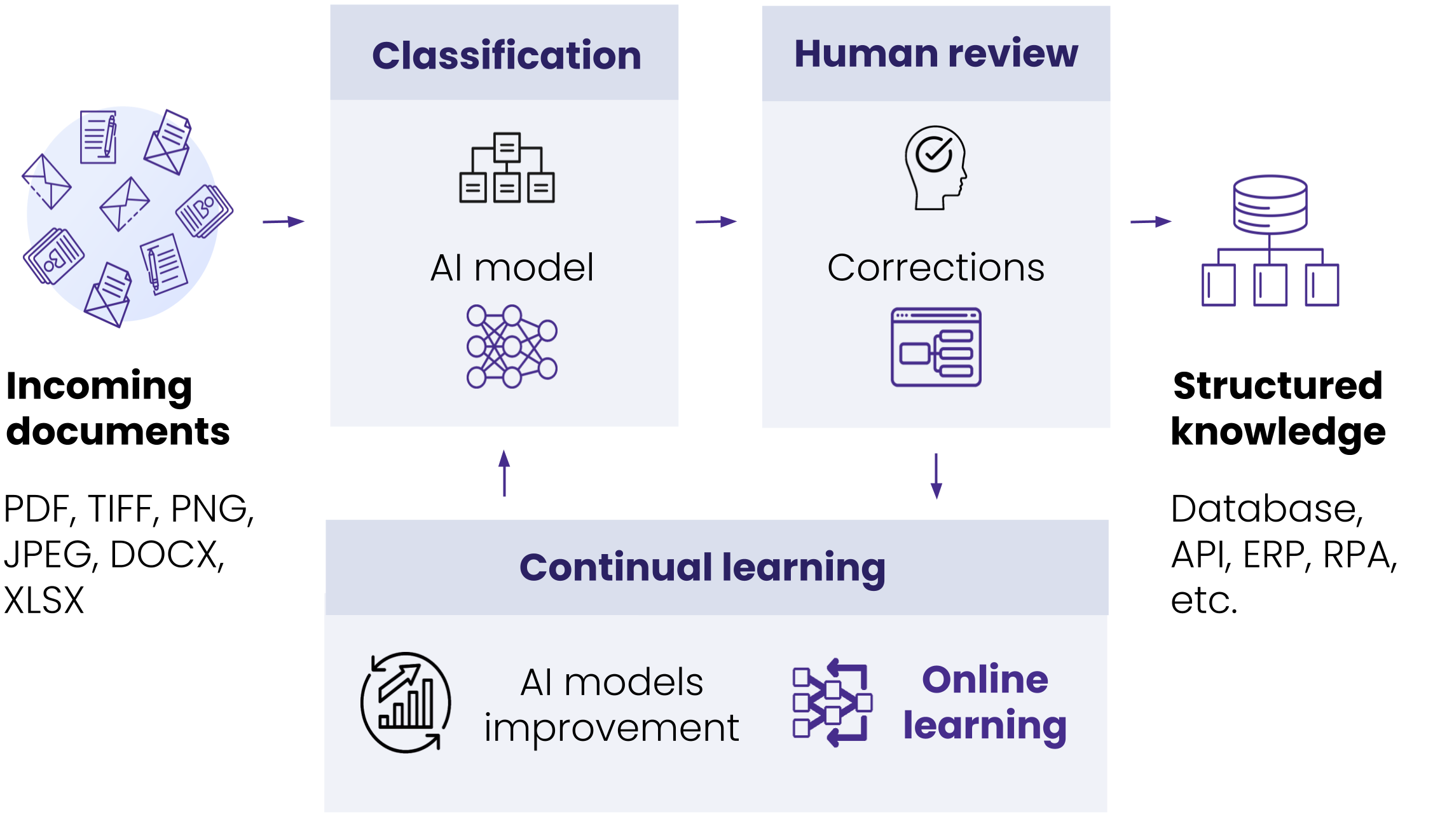}
  \captionof{figure}{Continual learning in document processing platform. Classification models need to learn incrementally and handle domain shifts after deployment.}
  \label{fig:problem-setup}
\end{figure}

Solutions based on deep neural networks have already found their applications in almost every domain that can be automated. An essential part of them is NLP, the development of which has gained particular momentum with the beginning of the era of transformers \cite{vaswani2017attention}. Complex and powerful models made it possible to solve problems such as text classification with a previously unattainable accuracy. However, exploiting the capabilities of such architectures in real-world systems requires online learning after deployment. This is especially difficult in dynamically changing environments that require the models to be frequently retrained due to domain or class setup shifts. An example of such environment is Alphamoon Workspace\footnote{https://alphamoon.ai/} where the presented architecture will be deployed as a model for document classification since we noticed the emerging need for online learning. We observed that the users' data in document classification process is changing frequently and such shifts often decrease the model accuracy. As a result, we have to retrain the models manually ensuing a time-consuming process. Our goal was to design an effective approach to incremental learning that will be used in a continual learning module of our system (Figure \ref{fig:problem-setup}).


Recently, neural architectures have become effective and widely used in classification problems \cite{devlin2018bert,rawat2017deep}. The parameter optimization process based on gradient descent works well when the data set is sufficiently large and fully available during the training process. Otherwise, the catastrophic forgetting \cite{french1999catastrophicforgetting} may occur, which makes neural networks unable to be trained incrementally. Continual learning aims to develop methods that enable accumulating new knowledge without forgetting previously learnt one. 

In this paper, we present a domain-agnostic architecture for online class incremental continual learning called DE\&E (Deep Encoders and Ensembles). Inspired by the E\&E method \cite{reference_method}, we proposed a method that increases its accuracy, provides full differentiability, and, most importantly, can effectively solve real-world classification problems in production environments. 
 Our contribution is as follows: 1) we introduced a differentiable KNN layer \cite{Diff-KNN-SOFT} into the model architecture, 2) we proposed a novel approach to aggregate classifier predictions in the ensemble, 3) we performed exhaustive experiments showing the ability to learn incrementally and real-world usability, 4) we demonstrate the effectiveness of the proposed architecture by achieving SOTA results on various data sets without a memory buffer.

\section{Related work}

\subsection{Continual Learning}

\subsubsection{Methods}

Currently, methods with a memory buffer such as GEM \cite{GEM}, A-GEM \cite{AGEM} or DER \cite{buzzega2020dark} usually achieve the highest performance in all continual learning scenarios \cite{RecencyBias}.
Such methods store part of the data in the memory and this data is successively replayed during training on new, unseen examples.
However, the requirement to store data in memory disqualifies these methods in many practical applications due to privacy policies or data size \cite{Safety1}. This forces attention toward other approaches, such as parameter regularization. The most popular methods in this group include EWC \cite{EWC} and LWF \cite{LWF}. When receiving a new dose of knowledge, these methods attempt to influence the model parameter updating procedure to be minimally invasive. 
As research shows \cite{3scenarios}, regularization-based methods fail in class incremental scenarios making them ineffective in many real-world cases. 

%

\begin{figure*}[t!]
\centering 
\includegraphics[width=1.0\linewidth]{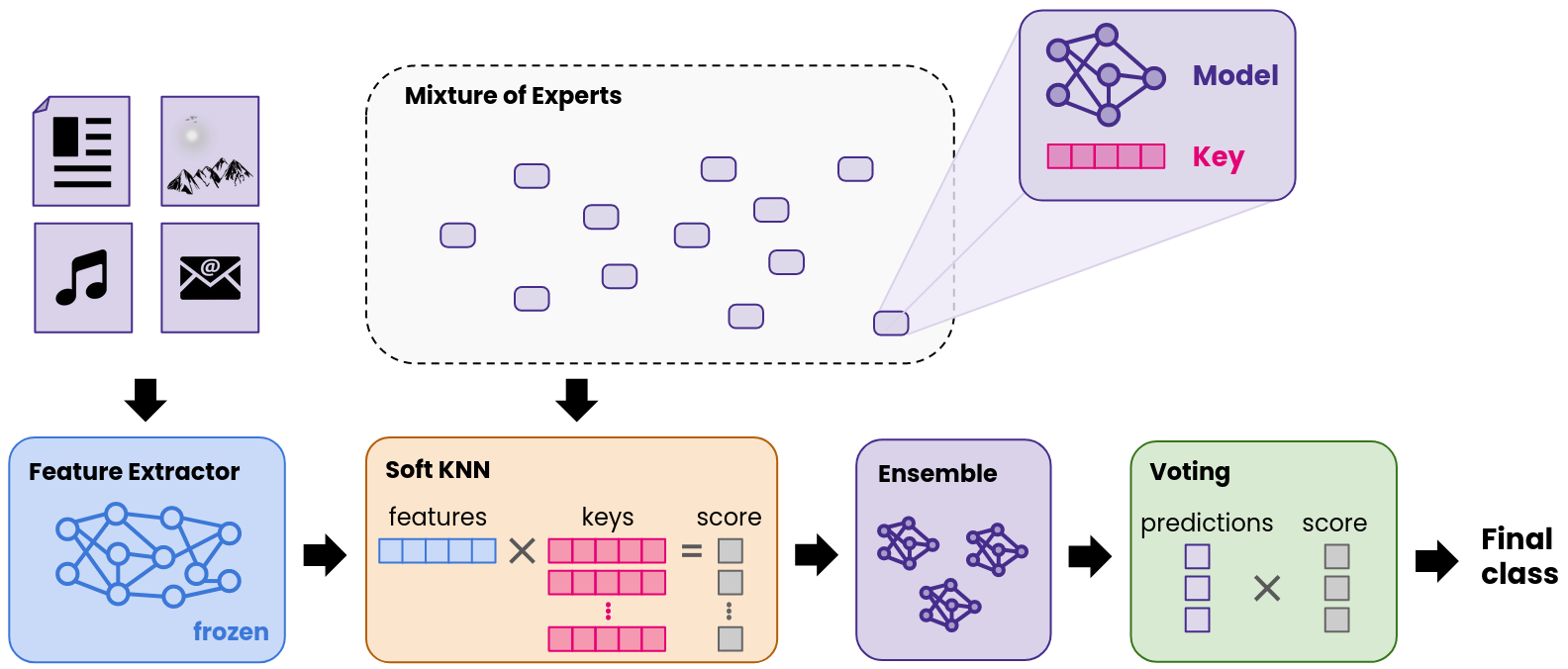}
\caption{Architecture of the proposed model. An input is processed by the feature extractor. Obtained embeddings are used to find the most relevant classifiers according to assigned keys. The \textit{soft KNN} layer approximates the \textit{soft KNN} scores. Predictions are weighted in the voting layer by both cosine similarity and \textit{soft KNN} scores. Final output is the class with the highest voting score.}
\label{fig:architecture_e2e}
\end{figure*}
 
\subsubsection{Approaches for NLP}
Almost all prior works focus on the development of continual learning methods in the computer vision domain \cite{delange2021continual}. Research on continual learning for NLP is limited and, as \citet{biesialska2020continual} observed, the majority of current
NLP methods are task-specific. Moreover, these methods often use a memory buffer \cite{de2019episodic} or relate to the language model itself \cite{ke2021achieving}. To address this niche, domain-agnostic approaches have to become much more prevalent in the near future.

\subsection{Ensemble methods}
Ensemble methods are widespread in the world of machine learning \cite{zhang2012ensemble}.
By using predictions of multiple weak learners, it is possible to get a model that performs surprisingly well overall. 
Broad adoption of methods \cite{cao2020ensemble, li2022novel, yang2021survey} demonstrates the effectiveness of ensemble techniques in a wide variety of tasks. Ensembles have also been used successfully in the field of continual learning, as evidenced by the BatchEnsemble \cite{wen2020batchensemble} or CN-DPM \cite{CN-DPM}. 
Other contributions present in literature \cite{doan2022efficient} tend to focus strongly on improving model performance rather than increasing model efficiency. Furthermore, ensemble approaches can also be used indirectly through dropout \cite{Dropout} or weights aggregation \cite{wortsman2022model}. 

\subsection{Mixture of Experts}
Mixture of Experts (ME) \cite{jacobs1991adaptive} is a technique based on the divide and conquer paradigm. It assumes dividing the problem space between several specialized models (experts). Experts are supervised by the gating network that selects them based on the defined strategy. The difference between the ensembles is that ME methods focus on selecting a few experts rather than combining predictions of all available models. ME techniques have found many applications in various domains \cite{masoudnia2014mixture}, including continual learning \cite{reference_method}, and even nowadays such approaches are widely used in NLP \cite{gao-etal-2022-parameter, ravaut-etal-2022-summareranker}. 

\subsection{Real-world NLP systems}
Over the last few years, the amount of real-world NLP applications has grown rapidly \cite{sarker2022ai}. Despite major successes in the real-world application of language technologies such as Google Translate, Amazon Alexa, and ChatGPT, production deployment and maintenance of such models still remain a challenge. Researchers have shown \cite{nowakowski-etal-2022-neyron, karakanta-etal-2021-simultaneous}, that there are several issues related to maintaining NLP models, including technical limitations, latency, and performance evaluation. However, the crucial problem is the shift of data domain that forces models to be retrained and deployed again over time \cite{hu2020challenges}. It is a major limitation in dynamically changing environments where users expect models to quickly adapt to them. Currently, this problem has been tackled in several systems \cite{afzal-etal-2019-development, hancock-etal-2019-learning}, but many of the solutions preclude maintaining model accuracy when training incrementally making them insufficient.

\section{Our approach} 
\subsection{Problem formulation}
Class incremental continual learning involves training a classification model $f(\cdot):\mathbb{X}\longmapsto\mathbb{Y}$ on a sequence of $T$ tasks. The model is trained on each task separately (one task at a time). Each task $D_{t}$ contains data points $D_{t}=\{(x^{1}_{t}, y^{1}_{t}), \dots, (x^{N_{t}}_{t}, y^{N_{t}}_{t})\}$, where $N_{t}$ is length of $D_{t}$, $x^{(i)}_{t}\in\mathbb{R}^{D}$, and $y^{(i)}_{t}\in\mathbb{Y}_{t}$. $\mathbb{Y}_{t}$ is a label set for task $t$ and $\mathbb{Y}_{t} \cap \mathbb{Y}_{t'} = \emptyset$ for $t \neq t'$.
We want the model to keep performing well on all previous tasks after each update, and we assume to be working in the most challenging setup \cite{3scenarios}, where one task consists of data from one class. 

\subsection{Method}

We present a flexible and effective domain-agnostic architecture that can be used to solve various classification problems. The architecture is presented in Figure \ref{fig:architecture_e2e}.

\paragraph{Feature extractor.}
The first component of the proposed architecture is a multi-layer feature extractor that transforms input data into the embedding space. It can be described by the following mapping $\mathbf{z}=F(\mathbf{x})$, where $\mathbf{x}\in\mathbb{R}^{D}$ is an input example and $\mathbf{z}\in\mathbb{R}^{M}$ is a $M$-dimensional embedding. The approach we follow assumes the use of a pre-trained model with frozen parameters. Such a procedure makes it possible to completely prevent the extractor from forgetting knowledge by isolating feature space learning from the classification process.

\paragraph{Keys and classifiers.}
We use an ensemble of $N$ classifiers $f_{n}(\cdot)$, where each of them maps the embedding into a $K$-dimensional output vector $\mathbf{\hat{y}}_{n}=f_{n}(\mathbf{z})$. With each classifier, there is an associated key vector $\mathbf{k}_{n}\in\mathbb{R}^{M}$ with the same dimensionality as the embedding. The keys help to select the most suitable models for specialization with respect to the currently processed input example. They are initialized randomly from normal distribution. We use simple single-layer neural networks as classifiers, with fan-in variance scaling as the weight initialization strategy. The network output is activated by a hyperbolic tangent function (\textit{tanh}).

\paragraph{Soft $\kappa$-nearest neighbors layer.}
The standard KNN algorithm is often implemented using ordinary sorting operations that make it impossible to determine the partial derivatives with respect to the input. It removes the ability to use KNN as part of end-to-end neural models. However, it is possible to obtain a differentiable approximation of the KNN model by solving the Optimal Transport Problem \cite{optimaltransportproblem}. Based on this concept, we add a differentiable layer to the model architecture. We call this layer soft $\kappa$-nearest neighbors (\textit{soft KNN}).
In order to determine the KNN approximation, we first compute a cosine distance vector $\mathbf{c}\in \mathbb{R}^{N}$ between the embedding and the keys:
\begin{equation}
    c_{n} = 1-\cos(\mathbf{z},\mathbf{k}_{n}),
\end{equation}
where $\mathbf{\cos(\cdot,\cdot)}$ denotes the cosine similarity. Next, we follow the idea of a soft top-$\kappa$ operator presented in \cite{Diff-KNN-SOFT}, where $\kappa$ denotes the number of nearest neighbors. Let $\mathbf{E}\in\mathbb{R}^{N\times 2}$ be the Euclidean distance matrix with the following elements:
\begin{equation}
e_{n,0}=(c_{n})^{2},\ \ \ e_{n,1}=(c_{n}-1)^{2}.    
\end{equation}
And let $\mathbf{G}\in\mathbb{R}^{N\times 2}$ denote the similarity matrix obtained by applying the Gaussian kernel to $\mathbf{E}$:
\begin{equation}
\mathbf{G}= \exp(-\mathbf{E}/\sigma),    
\end{equation} 
where $\sigma$ denotes the kernel width. The $\exp$ operators are applied elementwise to the matrix $\mathbf{E}$.

We then use the Bregman method, an algorithm designed to solve convex constraint optimization problems, to compute $L$ iterations of Bregman projections in order to approximate their stationary points:
\begin{equation}
    \mathbf{p}^{(l+1)}=\frac{\boldsymbol\mu}{\mathbf{G}\mathbf{q}^{(l)}},\ \ \ \mathbf{q}^{(l+1)}=\frac{\boldsymbol\nu}{\mathbf{G}^{\top}\mathbf{p}^{(l+1)}},\ \ \ 
\end{equation}
where $l=0,\dots,L-1, \boldsymbol\mu=\mathbf{1}_{N}/N$, $\boldsymbol\nu=[\kappa/N,(N-\kappa)/N]^{\top}$, $\mathbf{q}^{(0)}=\mathbf{1}_{2}/2$, and $\mathbf{1}_{i}$ denotes the $i$-element all-ones vector. 
Finally, let $\boldsymbol\Gamma$ denotes the optimal transport plan matrix and is given by:
\begin{equation}
\boldsymbol\Gamma = \mathrm{diag}(\mathbf{p}^{(L)})\cdot \mathbf{G} \cdot \mathrm{diag}(\mathbf{q}^{(L)})
\end{equation}
As the final result $\boldsymbol\gamma\in {\mathbb{R}^{N}}$ of the soft $\kappa$-nearest neighbor operator, we take the second column of $\boldsymbol\Gamma$ multiplied by $N$ i.e. $\boldsymbol\gamma=N  \boldsymbol\Gamma_{:,2}$. $\boldsymbol\gamma$ is a soft approximation of a zero-one vector that indicates which $\kappa$ out of $N$ instances are the nearest neighbors. Introducing the \textit{soft KNN} enables to train parts of the model that were frozen until now.

\paragraph{Voting layer.}
We use both $c_{n}$ and $\boldsymbol\gamma$ to weight the predictions by giving the higher impact for classifiers with keys similar to extracted features. The obtained approximation $\boldsymbol\gamma$ has two main functionalities. It eliminates the predictions from classifiers outside $\kappa$ nearest neighbors and weights the result. Since the Bregman method does not always completely converge, the vector $\kappa$ contains continuous values that are close to 1 for the most relevant classifiers. We make use of this property during the ensemble voting procedure. The higher the $\kappa$ value for a single classifier, the higher its contribution toward the final ensemble decision. The final prediction is obtained as follows:

\begin{equation}
\mathbf{\hat{y}}=\frac{\sum_{n=1}^{N} \gamma_{n}c_{n}\mathbf{\hat{y}}_{n}}{\sum_{n=1}^{N}c_{n}}    
\end{equation}

\paragraph{Training}
To effectively optimize the model parameters, we follow the training procedure presented in \cite{reference_method}. It assumes the use of a specific loss function that is the inner product between the ensemble prediction and the one-hot coded label: 

\begin{equation}
\mathcal{L}(\mathbf{y}, \hat{\mathbf{y}})=-\mathbf{y}^{\top} \hat{\mathbf{y}}
\end{equation}

Optimizing this criterion yields an advantage of using a \textit{tanh} activation function, significantly reducing catastrophic forgetting \cite{reference_method}. Following the reference method, we also use an optimizer that discards the value of the gradient and uses only its sign to determine the update direction. As a result, the parameters are being changed by a fixed step during the training.

\section{Experiments}


\subsection{Setup}

In order to ensure experiment's reproductivity, we evaluated our method on the popular and publicly available data sets.

\paragraph{Data sets}

\begin{table}[]
\caption{Data sets setup for experiments.}\label{tab:datasets}
\resizebox{\columnwidth}{!}{%
\begin{tabular}{llcccc}
\textbf{Domain}                           & \multicolumn{1}{c}{\textbf{Data set}} & \multicolumn{1}{c}{\textbf{Classes}} & \multicolumn{1}{c}{\textbf{Train}} & \multicolumn{1}{c}{\textbf{Test}} & \multicolumn{1}{c}{\textbf{Avg. words}} \\ \hline
\multicolumn{1}{c}{\multirow{3}{*}{Text}}                      & BBC News                           & 5                                    & 1,668                               &         557                          &      380                                   \\
 \multicolumn{1}{c}{} & Newsgroups                           & 10                                   &                       11314             &      7532                             &        315                                 \\
\multicolumn{1}{c}{}                      & Complaints                           & 10                                   &         16,000                           &      4,000                             &          228                               \\ \hline
\multicolumn{1}{c}{Audio}                                     & Speech Commands                      & 10                                   &          18,538                          &           2,567                        &                ---                         \\ \hline
\multicolumn{1}{c}{\multirow{2}{*}{Image}}                    & MNIST                                & 10                                   &   60,000                                 &             10,000                      &         ---                                \\
                                          & CIFAR-10                             & 10                                   &            50,000                        & 10,000                                  &                 ---                        \\ \hline
\end{tabular}%
}
\end{table}

\begin{table*}[]
\caption{Accuracy (\%) and standard deviation for methods evaluated on various data sets. Speech Commands data set was evaluated with 64 classifiers in ME, the remaining models have 128 classifiers. Regularization-based methods completely failed on the difficult data sets due to the recency bias phenomenon \cite{RecencyBias}.}\label{tab:experiments}
\begin{tabular*}{\textwidth}{lcllllllll}
\hline
    &  
    & \multicolumn{3}{c}{Text} 
    & \multicolumn{1}{c}{}          
    & \multicolumn{2}{c}{Image}  
    & \multicolumn{1}{c}{} 
    & \multicolumn{1}{c}{Audio} \\ 
    \cline{3-5} 
    \cline{7-8}  
    \cline{10-10}
\textbf{Model} & \textbf{Mem.} & \multicolumn{1}{c}{\textbf{NG}} & \multicolumn{1}{c}{\textbf{BBC}} & \multicolumn{1}{c}{\textbf{Compl.}} & \multicolumn{1}{c}{\textbf{}} & \multicolumn{1}{c}{\textbf{MNIST}} & \multicolumn{1}{c}{\textbf{CIFAR-10}} & \multicolumn{1}{c}{\textbf{}} & \multicolumn{1}{c}{\textbf{Sp. Comm.}} \\ \hline
Naive          & $\times$              & 5.25\tiny{$\pm$0.03}                                        & 21.65\tiny{$\pm$2.56}                                     & 9.56\tiny{$\pm$0.33}                                       &                               & 11.29\tiny{$\pm$3.05}                                 & 10.00\tiny{$\pm$0.01} & &          21.54\tiny{$\pm$3.78}                       \\
LwF            & $\times$              & 5.20\tiny{$\pm$0.05}                                        & 18.60\tiny{$\pm$2.03}                                     & 10.04\tiny{$\pm$0.20}                                       &                             & 11.47\tiny{$\pm$2.75}                                 & 10.00\tiny{$\pm$0.01} &  & 20.61\tiny{$\pm$3.88}                               \\
EWC            & $\times$              & 5.13\tiny{$\pm$0.13}                                        & 21.97\tiny{$\pm$2.14}                                     & 10.16\tiny{$\pm$0.31}                                       &                              & 11.19\tiny{$\pm$2.70}                                 & 10.00\tiny{$\pm$0.01} & &       32.93\tiny{$\pm$4.92}                            \\
SI             & $\times$              & 5.27\tiny{$\pm$0.01}                                        & 19.43\tiny{$\pm$2.96}                                     & 10.00\tiny{$\pm$0.62}                                       &                               & 14.90\tiny{$\pm$6.52}                                 & 10.00\tiny{$\pm$0.01} & &                9.99\tiny{$\pm$0.27}                   \\
CWR*           & $\times$              & 4.63\tiny{$\pm$0.60}                                        & 22.98\tiny{$\pm$1.20}                                     & 10.13\tiny{$\pm$0.33}                                       &                               & 10.40\tiny{$\pm$0.54}                                 & 10.00\tiny{$\pm$0.01} & &       10.32\tiny{$\pm$0.26}                           \\
GEM            & \checkmark             & 35.89\tiny{$\pm$3.80}                                        & 70.99\tiny{$\pm$7.68}                                     & 33.74\tiny{$\pm$2.50}                                       &                               & 52.27\tiny{$\pm$5.20}                                 & 23.40\tiny{$\pm$2.71} & &       21.01\tiny{$\pm$2.06}                           \\
A-GEM          & \checkmark             & 9.44\tiny{$\pm$7.14}                                        & 59.10\tiny{$\pm$17.52}                                     & 9.20\tiny{$\pm$0.01}                                       &                               & 65.37\tiny{$\pm$4.53}                                 & 26.43\tiny{$\pm$5.27} & &       17.45\tiny{$\pm$6.90}                           \\
Replay         & \checkmark             & 22.45\tiny{$\pm$3.09}                                        & 59.61\tiny{$\pm$3.17}                                     & 16.46\tiny{$\pm$4.62}                                      &                               & 69.02\tiny{$\pm$4.90}                                 & 32.93\tiny{$\pm$4.56} &  &       12.23\tiny{$\pm$1.28}                          \\ \hline
E\&E           & $\times$              & 46.07\tiny{$\pm$2.91}                                       & 75.87\tiny{$\pm$3.88}                                     & 44.80\tiny{$\pm$1.62}                                      &                               & 87.10\tiny{$\pm$0.21}                                 & 53.97\tiny{$\pm$1.31} & &  79.15\tiny{$\pm$0.60}                               \\
Ours          & $\times$              & \textbf{47.27}\tiny{$\pm$3.63}                                       & \textbf{78.49}\tiny{$\pm$3.92}                                    & \textbf{44.97}\tiny{$\pm$0.86}                                      &                               & \textbf{87.62}\tiny{$\pm$0.14}                                 & \textbf{56.27}\tiny{$\pm$1.21}   &    &  \textbf{80.11}\tiny{$\pm$1.30}                          \\ \hline
\end{tabular*}
\end{table*}

We use three common text classification data sets with different characteristics - Newsgroups \cite{Newsgroups20}, BBC News \cite{greene06icml}, and Consumer Finance Complaints\footnote{Source: \url{https://huggingface.co/datasets/consumer-finance-complaints}}. The goal of the experiments was to evaluate our method on tasks with with different difficulty levels. We also conducted experiments for audio classification using Speech Commands \cite{warden2018speech} data set. For the evaluation purposes, we selected the 10 most representative classes from the Newsgroups, Complaints and Speech Commands. Finally, we also conducted experiments on the popular MNIST and CIFAR-10 data sets as image domain representatives. The data set summary is presented in Table \ref{tab:datasets}. In all experiments we used a train set to train model incrementally, and afterward we performed a standard evaluation using a test set.

\paragraph{Feature extractors}

For all text data sets, we used a Distilbert \cite{sanh2019distilbert}, a light but still very effective alternative for large language models. Next, for Speech Commands, we utilized Pyannote \cite{Bredin2020}, a pretrained model for producing meaningful audio features. For image data sets, we used different extractors. MNIST features were produced by the pretrained VAE and CIFAR-10 has a dedicated BYOL model (see \ref{appendix:section:details} for more details).

\subsection{Results}

\begin{figure}
\centering
  \includegraphics[width=\linewidth]{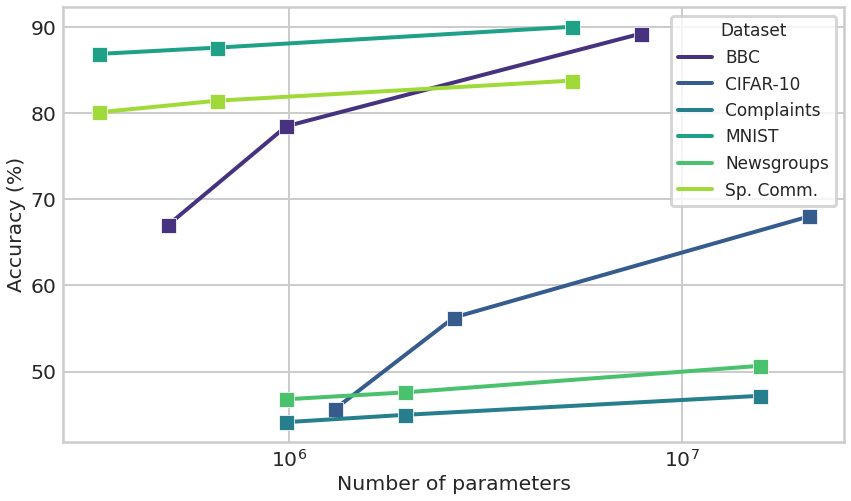}
  \captionof{figure}{Number of parameters in DE\&E architecture (64, 128, 1024 classifiers) and achieved accuracy (\%). We calculated the number of parameters as the sum of the parameters for all classifiers in the ME. Each mark is the test accuracy averaged across 5 runs.}
  \label{fig:parameters}
\end{figure}

The results of the evaluation are presented in Table \ref{tab:experiments}. For all setups evaluated, our model performed best improving results of the main reference method (E\&E) by up to 3 percent points (pp.). The improvement scale varies across the data sets. We also observed a significant difference in achieved accuracy between the DE\&E and the standard continual learning methods. Simple regularization-based methods completely fail in the class incremental scenario.
It shows how demanding training the model incrementally is when a set of classes is not fixed, which often takes place in real-world scenarios.
Furthermore, our method achieved these results without replaying training examples seen in the past, making it more practical relative to the SOTA memory-based methods (GEM, A-GEM, Replay) that store samples from every class. For the ensemble of 128 classifiers and Speech Commands data set, our architecture achieved an accuracy of more than 59 pp. higher than the best method with a memory buffer.

One of the most important hyperparameters of the model is the number of classifiers (experts). To investigate how it affects accuracy, we evaluated our architecture in three variants: small - 64, normal - 128, and large - 1024 classifiers. The evaluation results are presented in Figure \ref{fig:parameters}. We observed that increasing the ensemble size translates to higher accuracy, and gain depends on the setup and data characteristics. The most significant improvement was observed on BBC and CIFAR-10 where the large model achieved an accuracy of about 20pp. better than the small one. For the remaining data sets and the analogous setup, the gain was up to 5pp. We explain this phenomenon as the effect of insufficient specialization level achieved by smaller ensembles. If experts are forced to solve tasks that are too complicated they make mistakes often. Increasing the number of experts allows for dividing feature space into simpler sub-tasks. However, such a procedure has natural limitations related to the feature extractor. If features have low quality, increasing the number of experts will be ineffective. To select the optimal ensemble size we suggest using the elbow rule which prevents the model from being overparameterized and ensures reasonable accuracy. However, in general, we recommend choosing larger ensembles that are better suited for handling real-world cases.

Since real-world environments require deployed models to quickly adapt to domain shifts, we tested our method in a domain incremental scenario. In such setup, each data batch can provide examples from multiple classes that can be either known or new \cite{3scenarios}. This way, the model needs to learn incrementally, being prone to frequent domain shifts. As shown in Table \ref{tab:ci-di-comparison}, the proposed method handles both scenarios with comparable accuracy. We observed improved accuracy for BBC News, but reduced for the remaining data sets. Such property can be beneficial when there is limited prior knowledge about the data or the stream is imbalanced \cite{aguiar2022survey}.



\begin{table}[t]
\caption{Accuracy (\%) and standard deviation of DE\&E evaluated on Class Incremental and Domain Incremental scenarios. We used the same setup as shown in Table \ref{tab:experiments}.}
\label{tab:ci-di-comparison}
\resizebox{\columnwidth}{!}{%
\begin{tabular}{lcc}
 \multicolumn{1}{c}{\textbf{Data set}} & \multicolumn{1}{c}{\textbf{Class Incremental}} & \multicolumn{1}{c}{\textbf{Domain incremental}} \\ \hline
    BBC News        & 78.49\tiny{$\pm$3.92}   & 79.71\tiny{$\pm$3.14}   \\
    Newsgroups      & 47.27\tiny{$\pm$3.63}   & 44.55\tiny{$\pm$1.40}   \\
    Complaints      & 44.97\tiny{$\pm$0.86}   & 39.23\tiny{$\pm$3.03}   \\
    Speech Commands & 81.46\tiny{$\pm$0.85}   & 79.31\tiny{$\pm$0.49}   \\
    MNIST           & 87.62\tiny{$\pm$0.14}   & 85.04\tiny{$\pm$0.39}   \\ 
    CIFAR-10        & 56.27\tiny{$\pm$1.21}   & 55.66\tiny{$\pm$1.32}   \\ \hline
\end{tabular}}
\end{table}

\begin{figure}[h]
\centering
\includegraphics[width=\linewidth]{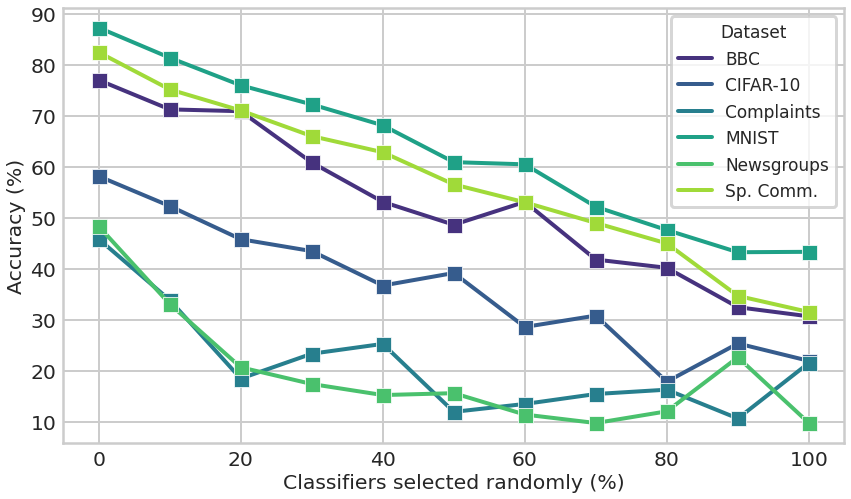}
\caption{Influence of random classifier selection on DE\&E accuracy (\%). All models consist of 128 classifiers. Each mark is the accuracy for an independent run.}
\label{fig:classifier-selection}
\end{figure}

We have also investigated the importance of the presented expert selection method. We trained the DE\&E method and for each training example, we allowed it to choose random experts (rather than the most relevant ones) with fixed probability $p$. As shown in Figure \ref{fig:classifier-selection}, the selection method has a strong influence on the model performance. Accuracy decreases proportionally to the $p$ over all data sets studied. The proper expert selection technique is crucial for the presented method. It is worth noting that relatively easier data sets suffer less from loss of accuracy than hard ones because even randomly selected experts can still classify the data by learning simple general patterns. In more difficult cases like Newsgroups and Complaints data sets, model performance is comparable to random guessing when $p > 0.5$.

\section{Conclusions}

In this paper, we proposed a domain-agnostic architecture for continual learning with a training procedure specialized in challenging class incremental problems. 
The presented architecture is based on the Mixture of Experts technique and handles many practical issues related to the deployment of text classification models in non-trivial real-world systems.
As our main contribution, we introduced a fully differentiable \textit{soft KNN} layer and a novel prediction weighting strategy.
By conducting exhaustive experiments, we showed improvement in accuracy for all the cases studied and achieved SOTA results without using a memory buffer. 
This enables an effective and secure training, especially when working with sensitive textual data. The presented architecture is highly flexible, can effectively solve classification problems in many domains, and can be applied to real-world machine learning systems requiring continuous improvement. Such work enables researchers to make further steps toward overrunning many of the current challenges related to language technology applications.

\section*{Limitations}
The main limitations of the proposed architecture are related to the presence of the frozen feature extractor.
The accuracy of the classification module is proportional to the quality of features. Since the ensemble weak learners are single-layer neural networks, the entire feature extraction process relies on a pre-trained model that strongly limits the upper bound of classification accuracy. Such approach reduces the method complexity, but also makes it prone to errors when embeddings have low quality. Achieving accuracy at a satisfactory level, which is crucial in real world systems, requires the use of high quality feature extractors. Currently, plenty of pretrained SOTA models are available for free in domains such as text or image classification, but if such extractor is not available, does not produce reasonable features or is too expensive to use, our architecture may not be the best choice.

Another issue is relatively long training time comparing to the reference methods (see \ref{appendix:time}). The introduction of a differentiable \textit{soft KNN} layer resulted in additional computational effort that clearly impacted the model complexity. This limits the use in low latency systems with machine learning models trained online.





\section*{Ethics Statement}

The authors foresee no ethical concerns with the work presented in this paper, in particular concerning any kind of harm and discrimination. Since the presented architecture can have a wide range of usages, the authors are not responsible for any unethical applications of this work.


\section*{Acknowledgements}
The research was conducted under the Implementation Doctorate programme of Polish Ministry of Science and Higher Education and also partially funded by Department of Artificial Intelligence, Wroclaw Tech and by the European Union under the Horizon Europe grant OMINO (grant number 101086321). It was also partially co-funded by the European Regional Development Fund within the Priority Axis 1 “Enterprises and innovation”, Measure 1.2. “Innovative enterprises, sub-measure 1.2.1. “Innovative enterprises – horizontal competition” as part of ROP WD 2014-2020, support contract no. RPDS.01.02.01-02-0063/20-00.



\bibliography{anthology,custom}
\bibliographystyle{acl_natbib}

\appendix

\section{Appendix}
\label{sec:appendix}

\subsection{Code}

Code is currently available as a Github repository \url{https://github.com/mateusz-wojcik-97/domain-agnostic-architecture}.

\subsection{Computing resources}

The machine we used had 128 GB RAM, an Intel Core i9-11900 CPU, and an NVIDIA GeForce RTX 3060 GPU with 12GB VRAM. Every experiment was performed using the GPU. 

\subsection{Time complexity}\label{appendix:time}

\begin{table}[h]
\caption{Time (seconds) of training the ensemble models with 128 classifiers on one task.}\label{tab:ensemble-time}
\centering
\begin{tabular}{lll}
\hline
\multicolumn{1}{c}{\textbf{Dataset}} & \multicolumn{1}{c}{\textbf{E\&E}} & \multicolumn{1}{c}{\textbf{Ours}} \\ \hline
Newsgroups & 7.43 & 31.20 \\
BBC News & 14.96 & 151.79 \\
Complaints & 20.33 & 93.63 \\
Sp. Commands & 30.80 & 108.90 \\
MNIST & 28.01 & 270.30 \\
CIFAR-10 & 104.25 & 355.82 \\ \hline
\end{tabular}
\end{table}

The comparison in training time between E\&E and DE\&E models is shown in Table \ref{tab:ensemble-time}. For all evaluated data sets, the training time of our model was higher than the time to train the reference method. The results vary between data sets. The introduction of a differentiable \textit{soft KNN} layer resulted in additional computational effort that clearly impacted the time complexity of the model.

\subsection{Implementation details}\label{appendix:section:details}

We use PyTorch to both reproduce the E\&E results and implement the DE\&E method. For text classification we used pretrained Distilbert \footnote{https://huggingface.co/distilbert-base-uncased} model and for audio classification we used pretrained Pyannote \footnote{https://huggingface.co/pyannote/embedding} model, both from the Huggingface repository. We used a pre-trained ResNet-50 model as the feature extractor for the CIFAR-10 data set. The model is available in the following GitHub repository, \url{https://github.com/yaox12/BYOL-PyTorch}, and is used under MIT Licence. For MNIST, we trained a variational autoencoder on the Omniglot data set and utilized encoder part as our feature extractor. We based our implementation of the \textit{soft KNN} layer on the code provided with \url{https://proceedings.neurips.cc/paper/2020/hash/ec24a54d62ce57ba93a531b460fa8d18-Abstract.html}. All data sets used are public.

\paragraph{Baselines}

\begin{table}[h]
    \caption{Architecture of neural networks used as backbones for baseline models depends on experimental setup. Each network has a similar number of total parameters as in the ensemble.}
    \label{tab:parameters-comparison}
    \centering
    \begin{tabular}{ll}
    \hline
    \textbf{Dataset} &         \textbf{Network layers}    \\ \hline
    Newsgroups & [1536, 1700, 768, 10]              \\
    Complaints & [1536, 955, 512, 10]              \\
    BBC News & [1536, 640, 5]              \\
    Sp. Commands & [512, 1256, 10]              \\
    MNIST & [512, 1256, 10]                  \\
    CIFAR-10 & [2048, 1274, 10]                   \\ \hline
    \end{tabular}
\end{table}

We use Naive, LwF \cite{LWF}, EWC \cite{EWC}, SI \cite{zenke2017continual}, CWR* \cite{lomonaco2017core50}, GEM \cite{GEM}, A-GEM \cite{AGEM} and Replay \cite{TinyEpisodicMemory} approaches as baselines to compare with our method. We utilize the implementation from Avalanche (\url{https://avalanche.continualai.org/}), a library designed for continual learning tasks. The main purpose of this comparison was to determine how the proposed method performs against classical approaches and, in particular, against the methods with memory buffer, which gives a significant advantage in class incremental problems. The recommended hyperparameters for each baseline method vary across usages in literature, so we chose them based on our own internal experiments. For a clarity, we keep hyperparameter naming nomenclature from the Avalnache library. For EWC we use $lambda$ = $10000$. The LwF model was trained with $alpha$ = $0.15$ and $temperature$ = $1.5$. For SI strategy, we use $lambda$ = $5e7$ and $eps$ = $1e-7$. The hyperparameters of the memory based approach GEM were set as follows: $memory\_strength$ = $0.5$, $patterns\_per\_exp$ = $5$, which implies that with every task, 5 examples will be accumulated. This has a particular importance when the number of classes is large. With this setup and 10 classes in data set, memory contains 50 examples after training on all tasks. Having a large memory buffer makes achieving high accuracy much easier. For the A-GEM method, use the same number of examples in memory and $sample\_size$ = $20$. All models were trained using Adam optimizer with a $learning\_rate$ of $0.0005$ and $batch\_size$ of $60$. We chose cross entropy as a loss function and performed one training epoch for each experience. To fairly compare baseline methods with ensembles, as a backbone we use neural network with a similar number of parameters (as in ensemble). Network architectures for each experimental setup are shown in Table \ref{tab:parameters-comparison}. All baseline models were trained by providing embeddings produced by feature extractor as an input.

\paragraph{Ensembles.}

We used E\&E \cite{reference_method} as the main reference method. It uses an architecture similar to that of a classifier ensemble, however the nearest neighbor selection mechanism itself is not a differentiable component and the weighting strategy is different. In order to reliably compare the performance, the experimental results of the reference method were fully reproduced. Both the reference method and the proposed method used exactly the same feature extractors. Thus, we ensured that the final performance is not affected by the varying quality of the extractor, but only depends on the solutions used in the model architecture and learning method.

Both E\&E and our DE\&E were trained with the same set of hyperparameters (excluding hyperparameters in the \textit{soft KNN} layer for the DE\&E). We use ensembles of sizes 64, 128 and 1024. Based on the data set, we used different hyperparameter sets for the ensembles (Table \ref{tab:ensemble-hp}).

The keys for classifiers in ensembles were randomly chosen from the standard normal distribution and normalized using the $L2$ norm. The same normalization was applied to encoded inputs during lookup for matching keys. 

\begin{table*}[!h]
\caption{Hyperparameters used for DE\&E and E\&E methods.}\label{tab:ensemble-hp}
\begin{tabular*}{\textwidth}{c|l|l|l|l|l}
\hline
\textbf{Dataset} & \multicolumn{1}{c|}{\textbf{Classifiers}} & \multicolumn{1}{c|}{\textbf{Neighbors}} & \multicolumn{1}{c|}{\textbf{Batch size}} & \multicolumn{1}{c|}{\textbf{Learning rate}} & \textbf{Weight Decay} \\ \hline
\multirow{3}{*}{Newsgroups} & 64 & 16 & \multirow{3}{*}{8} & \multirow{3}{*}{0.0001} & \multirow{14}{*}{0.0001} \\ \cline{2-3}
 & 128 & 32 &  &  &  \\ \cline{2-3}
 & 1024 & 64 &  &  &  \\ \cline{1-5}
\multirow{3}{*}{BBC News} & 64 & 8 & \multirow{3}{*}{1} & \multirow{3}{*}{0.01} &  \\ \cline{2-3}
 & 128 & 16 &  &  &  \\ \cline{2-3}
 & 1024 & 32 &  &  &  \\ \cline{1-5}
\multirow{3}{*}{Complaints} & 64 & 16 & \multirow{3}{*}{8} & \multirow{3}{*}{0.0001} &  \\ \cline{2-3}
 & 128 & 32 &  &  &  \\ \cline{2-3}
 & 1024 & 64 &  &  &  \\ \cline{1-5}
\multirow{3}{*}{Sp. Commands} & 64 & 16 & \multirow{3}{*}{8} & \multirow{3}{*}{0.001} &  \\ \cline{2-3}
 & 128 & 32 &  &  &  \\ \cline{2-3}
 & 1024 & 64 &  &  &  \\ \cline{1-5}
MNIST & 128 & 16 & 60 & 0.0001 &  \\ \cline{1-5}
CIFAR-10 & 128 & 16 & 60 & 0.0001 &  \\ \hline
\end{tabular*}
\end{table*}

\paragraph{Soft KNN.} 

We use the Sinkhorn algorithm to perform the forward inference in \textit{soft KNN}. The Sinkhorn algorithm is useful in entropy-regularized optimal transport problems thanks to its computational effort reduction. The \textit{soft KNN} has $\mathcal{O}(n)$ complexity, making it scalable and allows us to safely apply it to more computationally expensive problems. 

The values of \textit{soft KNN} hyperparameters were $\sigma = 0.0005$ and $L = 400$. We utilize the continuous character of an output vector to weight the ensemble predictions. It is worth noting that we additionally set the threshold of the minimum allowed \textit{soft KNN} score to 0.3. It means every element in $\boldsymbol\gamma$ lower than 0.3 is reduced to 0. We reject such elements because they are mostly the result of non-converged optimization and do not carry important information. In this way, we additionally secure the optimization result to be as representative as possible.

\end{document}